\pdfoutput=1

\documentclass[11pt]{article}

\usepackage{EACL2023}

\usepackage{times}
\usepackage{latexsym}

\usepackage[T1]{fontenc}

\usepackage[utf8]{inputenc}

\usepackage{microtype}

\usepackage{inconsolata}

\usepackage{comment}
\usepackage{import}
\usepackage{graphicx}
\usepackage{multirow}
\usepackage{booktabs}

\title{Global Constraints with Prompting for\\ Zero-Shot Event Argument Classification}

\author{Zizheng Lin$^{1}$, Hongming Zhang$^{2}$ \& Yangqiu Song$^{1}$ \\
  $^{1}$Department of Computer Science and Engineering, HKUST \\
  $^{2}$Tencent AI Lab, Seattle \\
  \texttt{\{zlinai, yqsong\}@cse.ust.hk}; \texttt{hongmzhang@global.tencent.com}}

\begin{document}
\maketitle
\begin{abstract}
Determining the role of event arguments is a crucial subtask of event extraction.
Most previous supervised models leverage costly annotations, which is not practical for open-domain applications.
In this work, we propose to use global constraints with prompting to effectively tackles event argument classification without any annotation and task-specific training.
Specifically, given an event and its associated passage, the model first creates several new passages by prefix prompts and cloze prompts, where prefix prompts indicate event type and trigger span, and cloze prompts connect each candidate role with the target argument span. Then, a pre-trained language model scores the new passages, making the initial prediction.
Our novel prompt templates can easily adapt to all events and argument types without manual effort. Next, the model regularizes the prediction by global constraints exploiting cross-task, cross-argument, and cross-event relations. 
Extensive experiments demonstrate our model's effectiveness: it outperforms the best zero-shot baselines by 12.5\% and 10.9\% F1 on ACE and ERE with given argument spans and by 4.3\% and 3.3\% F1, respectively, without given argument spans.
We have made our code publicly available.\footnote{https://github.com/HKUST-KnowComp/Constraints-with-Prompting-for-Zero-Shot-EAC}
\end{abstract}

\section{Introduction}
Event Argument Classification\footnote{We focus on event argument because existing zero-shot trigger extraction models like \citet{ZhangWR21} are already strong enough, but the arguments remain a challenge. Our argument identification approach is described in Section \ref{sec:settings}.} (EAC), finding the roles of event arguments, is an important and challenging event extraction sub-task. 
\begin{figure}
    \centering
    \resizebox{0.85\columnwidth}{!}{
    \includegraphics{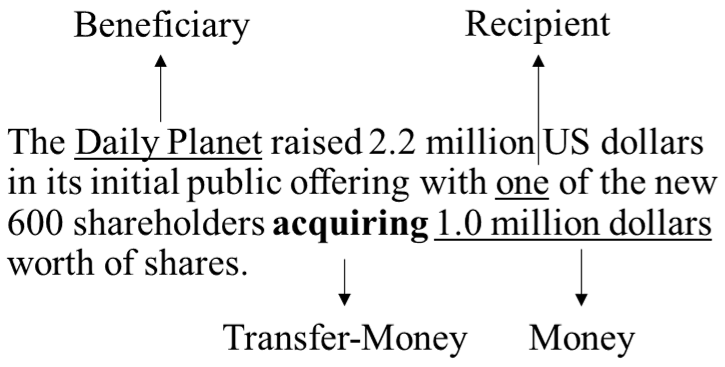}
    }
    \caption{An example of EAC. The trigger is in \textbf{bold face}. Arguments are \underline{underlined} and connected to their roles by arrows.}
    \label{fig:eac_example}
\end{figure}
As shown in Figure \ref{fig:eac_example}, a ``Transfer-Money'' event whose trigger is ``acquiring'' has several argument spans (e.g., ``Daily Planet''). By determining the role of these arguments (e.g., ``Daily Planet'' as ``Beneficiary''), we can obtain a better understanding of the event, thus benefiting related applications like stock price prediction \citep{DingZLD15} and biomedical research \citep{ZhaoZYHML21}.

Many previous EAC works require numerous annotations to train their models \citep{LinJHW20,naacl2022degree,LiuHSW22}, which is not only costly as the annotations are labor-intensive but also difficult to be generalized to datasets of novel domains.
Accordingly, some EAC models adopt a few-shot learning paradigm \citep{MaW0LCWS22,naacl2022degree}. However, they are sensitive to the few-shot example selection and they still require costly task-specific training, which hinders their real-life deployment.
There have been some zero-shot EAC models based on transfer learning \citep{DaganJVHCR18}, or label semantics \citep{ZhangWR21, WangYCSH22}, or prompt learning \citep{LiuCLBL20,LyuZSR20,HuangHNCP22,abs-2204-02531}. 
However, these models' corresponding limitations impede their real-life deployment. The model based on transfer learning can be ineffective when new event types are very different from the observed one. As for models using label semantics, they require a laborious preparation process and have unsatisfactory performance. Regarding models adopting prompt learning, they need tedious prompt design customized to every new type of events and arguments, and their performance is also limited.

To address the aforementioned issues, we propose an approach using global constraints with prompting to tackle zero-shot EAC. 
Global constraints can be viewed as a type of supervision signal from domain knowledge, which is crucial for zero-shot EAC since supervision from annotations is inaccessible. Moreover, our model's constraints module provides abundant global insights across tasks, arguments, and events.
Prompting can also be regarded as a supervision signal as it induces abundant knowledge from Pre-Trained Language Models (PTLM). 
Unlike previous zero-shot EAC works, which need a tedious prompt design for every new type of events and arguments, the novel prompt templates of our model's prompting module can be easily adapted to all possible types of events and arguments in a fully automatic way.
Specifically, given an event and its passage, our model first adds prefix prompt, cloze prompt, and candidate roles into the passage, which creates a set of new passages. The Prefix prompt describes the event type and trigger span. Cloze prompt connects each candidate to the target argument span. Afterwards, our model adopts a PTLM to compute the language modeling loss for each of the new passages, whose negative value would be the respective prompting score.
The role with the highest prompting score is the initial prediction.
Then, our model uses global constraints to regularize the initial prediction. The global constraints are based on the domain knowledge of the following relations: (1) cross-task relation, where our model additionally performs another one or more classification task on target argument span, and our model's predictions on EAC and other task(s) should be consistent; (2) cross-argument relation, where arguments of one event should collectively abide by certain constraint(s); (3) cross-event relation, where some argument playing a certain role in one event should play a typical role in another related event.

We conduct comprehensive experiments to demonstrate the effectiveness of our model. Particularly, our approach surpasses all zero-shot baselines by at least 12.5\% and 10.9\% F1 on ACE and ERE, respectively. When argument spans are not given, our model outperforms the best zero-shot baseline by 4.3\% and 3.3\% F1 on ACE and ERE, respectively. 
Besides that, we also conduct experiments to show that both the prompting and constraints modules contribute to the final success.

\section{Methodology}
We first present an overview of our approach. Then we introduce the details by describing its prompting module and global constraints regularization module.
We follow \citep{DBLP:journals/corr/abs-2107-13586} to name a prompt inserted before input text as \textit{prefix prompt}, and a prompt with slot(s) to fill in and insert in the middle of input text as \textit{cloze prompt}.

\subsection{Overview}\label{sec:model_overview}
\begin{figure*}[t]
    \centering
    \resizebox{2\columnwidth}{150pt}{
    \includegraphics{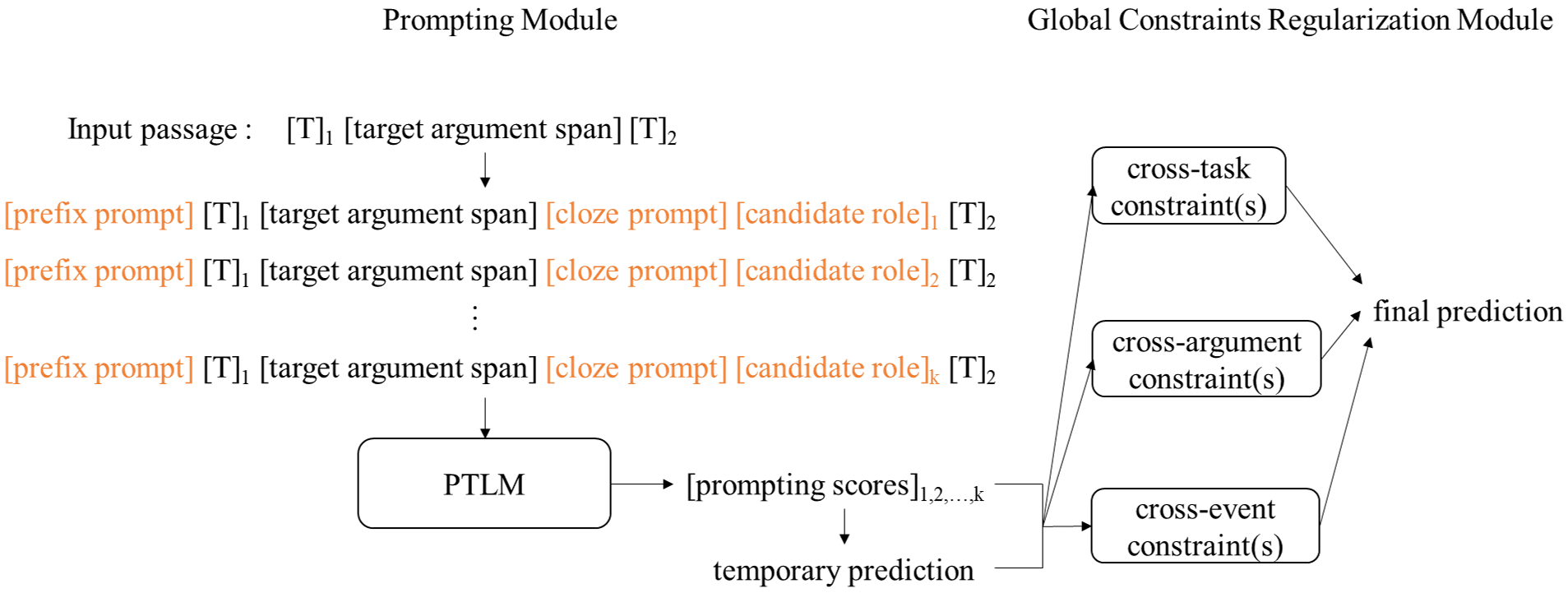}
    }
    \caption{Model overview using prediction for one argument span as an example. $[T]_1$ and $[T]_2$ are the parts of the input passage before and after the span, respectively. $k$ is the number of candidate roles of the event type.}
    \label{fig:model_overview}
\end{figure*}
As shown in Figure \ref{fig:model_overview}, given a passage with a target argument span, our model infers the target's role without annotation and task-specific training. Our model has two modules.
The first module is the prompting module that creates and scores several new passages. During creation, the model adds prefix prompt, cloze prompt, and candidate roles into the passage, where the prefix prompt contains information about event type and trigger, and the cloze prompt joins each candidate with a target argument span.\footnote{Since we focus on event argument classification, we assume that the event types and trigger spans are given. The settings without given argument spans will be discussed in Section \ref{sec:settings}.} Afterwards, the model uses a PTLM to score the new passages. 
Our novel prompt templates can easily adapt to all possible events and arguments without manual work.
Initial prediction is the role with the best prompting scores.
The second module is the global constraints regularization module, where the model regularizes the prediction by three types of global constraints: cross-task constraint, cross-argument constraint, and cross-event constraint. All global constraints are based on event-related domain knowledge about inter-task, inter-argument, and inter-event relations.

\subsection{Prompting Module}\label{sec:prompting_module}
In this section, we describe the prompting module in detail.
Given a passage, we first add a prefix prompt containing information about the event type and trigger span to the beginning. Such a prompt can guide a PTLM to: (1) accurately capture the input text's perspective related to the event; (2) have a clear awareness of the trigger. 
Based on the definitions of events and triggers \citep{grishman2005nyu}, we create the following prefix prompt: ``\textbf{This is a [] event whose occurrence is most clearly expressed by [].}'' where the first and second pairs of square brackets are the placeholders of event type and trigger span respectively. We also conducted some experiments comparing different prefix prompts in Section \ref{sec:prefix_comp}, and the results showed that the prefix above is the most effective.

Second, for each candidate role, the module inserts the cloze prompt behind the target argument span, and the role fills the prompt' slot. 
The cloze prompt adopts the hypernym extraction pattern ``\textbf{M and any other []}'' \citep{DaiSW20}, where ``M'' denotes the argument span and the square bracket is the placeholder of the candidate role. We did not try other hypernym extraction patterns as \citep{DaiSW20} had shown that our pattern is the most effective.
The motivation for adopting the hypernym extraction pattern for cloze prompt is that, to some extent a role can be regarded as a context-specific hypernym of the respective argument span of the associated event (e.g., ``Beneficiary'' can be seen as a context-specific hypernym of ``Daily Planet''of the Transfer-Money event described by the example in Figure \ref{fig:eac_example}). Hence, such a prompt induces the linguistic and commonsense knowledge stored in PTLM to help identify which candidate role is the most reasonable. 

After adding the previous two types of prompts, we get several new passages. For instance, suppose the passage is``In Baghdad, a bomb was fired at 17 people.'' whose event type is ``Conflict:Attack'', trigger is ``fired'', target argument span is ``bomb'', and candidate roles are \{``Attacker'', ``Instrument'', ``Place'', ``Time'', ``Target''\}. 
The created passages would be: (1) ``\textit{This is a Attack event whose occurrence is most clearly expressed by ``fired.''} In Baghdad, a bomb \textit{and any other \underline{attacker}} was fired at 17 people.''; (2) ``\textit{This is a Attack ... ``fired.''} ... bomb \textit{and any other \underline{instrument}} was ...''; and similar text for other roles.\footnote{We only use the subtype of all events following the pre-processing done by \citep{LinJHW20}}

For each new passage, we apply a PTLM to compute the language modeling loss.
The negative value of the loss would be the prompting score of the respective passage, where a higher value indicates higher plausibility according to the PTLM.
\textbf{Since our model's prompt templates are independent of event type and argument role, their adaptation to any new type of events and arguments is trivial and fully automatic.} Hence, our prompting method is more scalable and generalizable than those of previous zero-shot EAC models, since, for every new type of events and arguments they need to design a customized prompt. For instance, for every type of events/arguments, \citet{LyuZSR20} manually design a unique prompt as text entailment/question answering template.
The initial prediction would be the role with the highest prompting score.
Since the steps of obtaining scores for each candidate role are independent of other candidate roles, we implement the steps of different candidate roles in parallel. Such a parallel implementation significantly improves our model's efficiency.

\subsection{Global Constraints Regularization Module}\label{sec:constraints}
This module regularizes the prediction by the following three types of global constraints.\footnote{We designed 14 global constraints in total and we used preliminary experiments to choose the three most effective ones. In the preliminary experiments, we randomly sample 1k instances covering all trigger and argument types. We then evaluate each constraint on the sampled subset.}

\begin{figure}[t]
    \centering
    \resizebox{0.8\columnwidth}{!}{
    \includegraphics{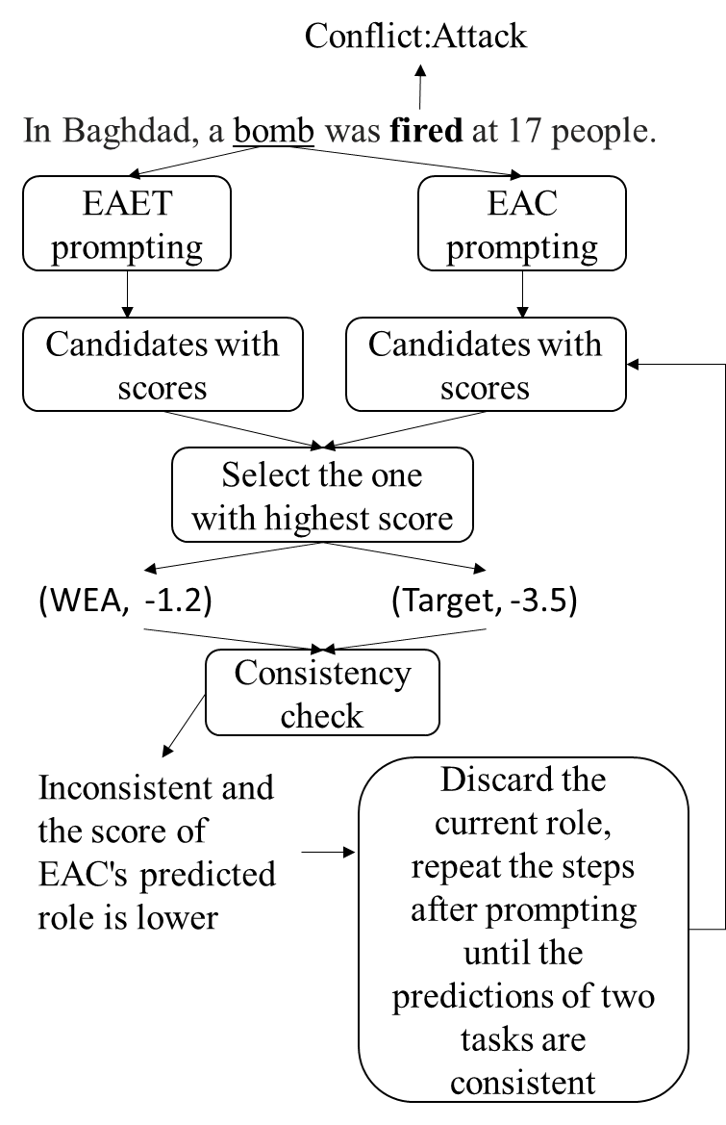}
    }
    \caption{An Example of cross-task constraint. The text in \textbf{bold face} is the trigger, \underline{underlined text} is target argument span, and a tuple denotes a predicted label with its prompting score (e.g.,  ``(Target, -3.5)’’ denotes the predicted label ``Target’’ with its prompting score``-3.5’’). Similar notations are adopted in all remaining figures.}
    \label{fig:cross-task_constraint}
\end{figure}

\textbf{Cross-task constraint} exploits the label dependency between EAC and auxiliary task(s) so that our model can get global information from the auxiliary task(s) about event arguments. We use \textbf{Event Argument Entity Typing (EAET)} as the auxiliary task. The task aims to classify an argument into its context-dependent entity type (e.g., PER). \textbf{As specified in ACE2005 ontology, an argument of a certain role in an event can only be one of several respective entity types (e.g., an argument of ``Attack'' role in a Conflict:Attack event can only be ``ORG,'' ``PER,'' or ``GPE'').} Based on this domain knowledge, we design the cross-task constraint as follows: (1) For each input passage, our model performs prompting for EAET, where the prompting is the same as in Section \ref{sec:prompting_module} except that candidate entity types replace the candidate roles in cloze prompt.; (2) After obtaining the scores and prediction of EAET, the model check the consistency between the predictions of EAC and EAET; (3) If the consistency is violated and the score of EAC's predicted role is lower, then discard the current role, use the role with the highest score in the remaining ones, and check the consistency again; (4) The constraint ends when the labels of two tasks are consistent. An example illustrating this type of constraint is shown in Figure \ref{fig:cross-task_constraint}.

\begin{figure}[t]
    \centering
    \resizebox{0.6\columnwidth}{!}{
    \includegraphics{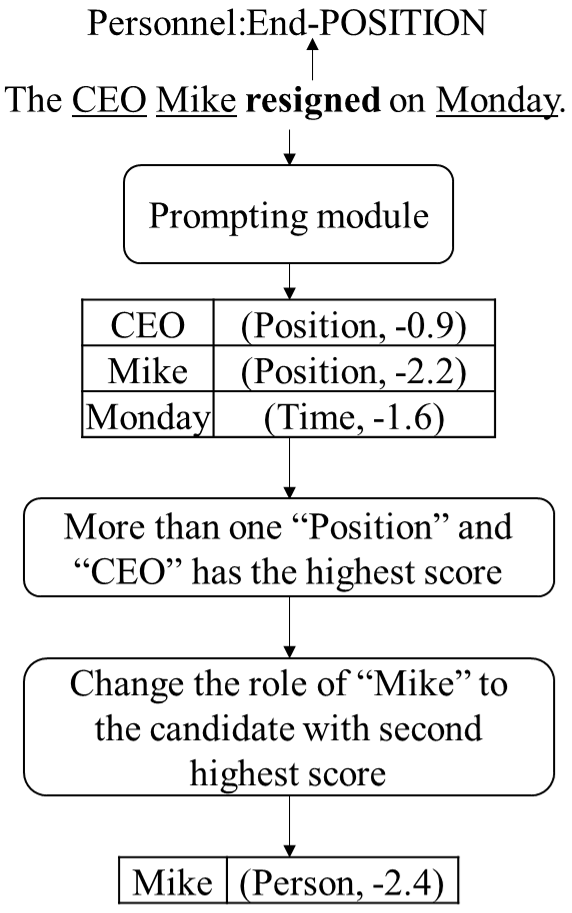}
    }
    \caption{An Example of cross-argument constraint.}
    \label{fig:cross-argument_constraint}
\end{figure}

\textbf{Cross-argument constraint} is based on domain knowledge about relationships between arguments within an event. Specifically, our model constrains the number of particular arguments for some or all events. For instance, it is very unlikely that an event mentioned is associated with multiple ``Time'' arguments. Such constraints offer a global understanding of event arguments to our model. The cross-argument constraint we adopt is ``\textbf{A Personnel:End-POSITION event has at most one Position argument}.'' Given a Personnel:End-POSITION event, our model first checks the number of ``Position'' argument. If the number is more than one, then our model will first collect the arguments whose roles are ``Position'' and remove the one with the highest score among these arguments. Then for each remaining argument, our model would change the role to its candidate with the second highest score. An example illustrating this type of constraint is shown in Figure \ref{fig:cross-argument_constraint}.

\begin{figure}[t]
    \centering
    \resizebox{\columnwidth}{!}{
    \includegraphics{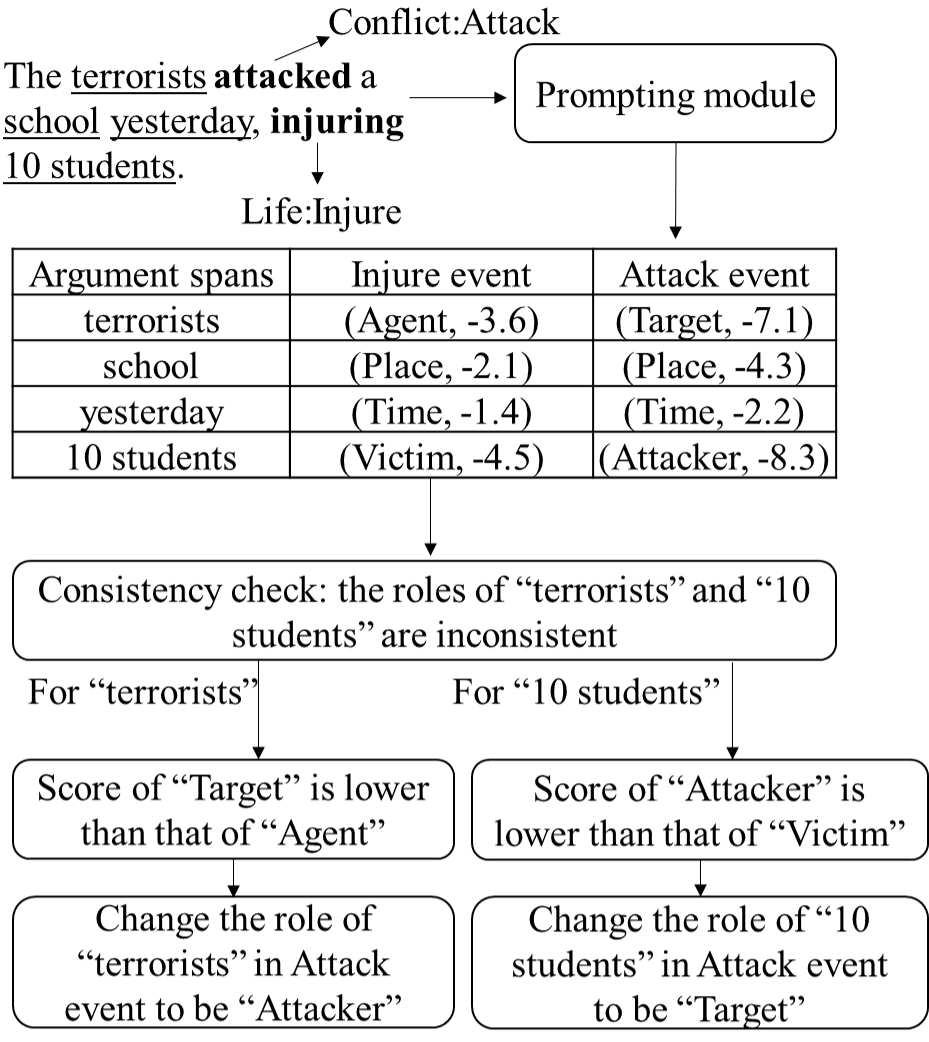}
    }
    \caption{An Example of cross-event constraint.}
    \label{fig:cross-event_constraint}
\end{figure}

\textbf{Cross-event constraint} regularizes predicted roles of arguments shared by related events. A model with such a constraint can have global insights into event arguments, because while they are making inferences for the arguments of one event, they are aware of the information of other related event(s) and cross-event relations.
The cross-event constraint we adopt is ``\textbf{If a Life:Injure event and a Conflict:Attack event share arguments, then Injure.Place is the same as Attack.Place, Injure.Victim is the same as Attack.Target, Injure.Instrument is the same as Attack.Instrument, Injure.Time is the same as Attack.Time, Injure.Agent is the same as Attack.Attacker}''. Given a passage containing an Injure and an Attack event sharing arguments, the model imposes the constraint by checking the consistency between the respective roles of each shared argument as specified in the constraint. Any inconsistency would be fixed by changing the role with a lower prompting score to the new one satisfying the consistency. An example illustrating this type of constraint is shown in Figure \ref{fig:cross-event_constraint}.

Our constraint modeling method can be easily generalized to other datasets/ontologies by simply using the knowledge about corresponding cross-task, cross-argument, and cross-event relations to design new constraints. The design processes are not costly as we could easily find such knowledge from the guidelines of the target dataset.

\section{Experiments}
We first present the experimental settings, baselines used for comparison, and some implementation details. Next, we show and analyze the experiment results. Then we present a detailed analysis of the prompting module and global constraints regularization module. Finally, we conduct an error analysis.

\subsection{Settings}\label{sec:settings}
We use ACE (2005-E$^+$)\footnote{https://www.ldc.upenn.edu/collaborations/past-projects/ace} \citep{DoddingtonMPRSW04,LinJHW20} and ERE(-EN) \citep{SongBSRMEWKRM15} as datasets. In total, ACE has 33 event types and 22 roles, whereas ERE has 38 event types and 21 roles. We pre-process all events to keep only the event subtypes whenever applicable, as done in \citep{LinJHW20}.
Following the pre-processing in \citep{ZhangWR21}, for each dataset, we merge all splits into one test set since our approach is zero-shot.
When argument spans are not given, we pipeline our model with an argument identification module adapted from \citep{LyuZSR20}. Specifically, we replace the QA model in \citep{LyuZSR20} with a more powerful PTLM with a span classification head on top, and the whole model has been fined-tuned for extractive QA tasks. Then for a passage, we prompt each role using the new QA model as in \citep{LyuZSR20}. We collect the prompt results for all roles (ignoring the ``None'' result) as candidate spans for the passage.
We use the F1 score for evaluation following \citep{JiG08}, where argument spans are evaluated on the head level when not given.
Regarding PTLMs, We use GPT-J (6B) \citep{gpt-j} instances from Huggingface \citep{WolfDSCDMCRLFDS20}, where an instance for causal language modeling is used for prompting, and an instance for QA is used for argument identification.
In all the following sections except Section \ref{sec:main_res}, we conduct experiments on ACE, assuming that argument spans are given.

\subsection{Main Results}\label{sec:main_res}
\begin{table*}[!ht]
    \centering
    \resizebox{2\columnwidth}{!}{
    \begin{tabular}{ccccc}
    \toprule
        \multirow{2}{*}{Model} & \multicolumn{2}{c}{ACE} & \multicolumn{2}{c}{ERE} \\
        & argument span given & argument span not given & argument span given & argument span not given \\ 
        \midrule
        \citep{naacl2022degree} (supervised) & 79.3 & 71.8 & 79.8 & 72.5 \\ 
        \midrule
        \citep{LiuCLBL20} & 46.1 & 24.2 & 40.9 & 22.8 \\ 
        \citep{LyuZSR20} & 47.8 & 26.9 & 44.5 & 26.3 \\ 
        \citep{ZhangWR21}  & 53.6 & 23.5 & 51.9 & 20.2 \\ 
        Ours & \textbf{66.1} & \textbf{31.2} & \textbf{62.8} & \textbf{29.6} \\ \bottomrule
    \end{tabular}
    }
    \caption{Performance of supervised model, zero-shot baselines, and our model. The best scores among the ones of zero-shot methods are in bold font.}
    \label{tab:main_res}
\end{table*}
We report the main results comparing our models with three previous powerful zero-shot models \citep{LiuCLBL20,LyuZSR20,ZhangWR21}. Moreover, we also report the results of a SOTA supervised model \citep{naacl2022degree}. We obtain the results of all compared methods from our own experiments to ensure a fair comparison on the same datasets and same settings.
From Table \ref{tab:main_res}, we have the following observations:
\begin{itemize}
    \item Our model achieves superior performance on both datasets under both settings compared with all zero-shot baselines. Specifically, our model surpasses the best zero-shot baselines \citep{ZhangWR21} by 12.5\% and 10.9\% on ACE and ERE, respectively. Without argument spans, our model outperforms the respective best zero-shot baselines \citep{LyuZSR20} by 4.3\% and 3.3\% on ACE and ERE, respectively, which is also a noticeable gap. Such large performance improvements can be attributed to the following: (1) the prefix prompt guides the PTLM to effectively capture input's event-related perspective and trigger; (2) the cloze prompt leverages linguistic and commonsense knowledge stored in PTLM to improve its contextual understanding of event arguments; (3) the global constraints regularization incorporate global information and domain knowledge in inference. In Section \ref{sec:analyze_prompt}, we compare the effects of using different PTLMs like BERT in the prompting module, and the results show that our model consistently outperforms previous zero-shot models, as shown in Table \ref{tab:main_res} and Figure \ref{fig:analysis_prompt_ptlm}.
    \item Compared with the supervised SOTA model \citep{naacl2022degree}, there is still a significant gap between our model's performance and that it. Specifically, \citep{naacl2022degree} outperforms our model by 13.2\% and 17.0\% on ACE and ERE, respectively. When argument spans are not provided, \citep{naacl2022degree} outruns our model by 40.6\% and 42.9\% on ACE and ERE, respectively. We can see that the advantage of supervised SOTA over our zero-shot method is much more distinct when argument spans are not given in advance. This is probably because our zero-shot argument identification module described in Section \ref{sec:settings} is not powerful enough, which causes severe error propagation to our EAC model.
\end{itemize}

\subsection{Analysis of Prompting Module}\label{sec:analyze_prompt}
We conduct experiments to examine the effects of different configurations of prefix prompt templates. 
Specifically, we compare our model's complete prefix prompt with the following configurations: (1) removing event type information from the prefix; (2) removing trigger information from the prefix; (3) removing the whole prefix.
For instance, suppose the passage is ``In Baghdad, a bomb was fired at 17 people.'' mentioned in Section \ref{sec:prompting_module}, the prefix in configuration (1) would be \textbf{``This event's occurrence is most clearly expressed by `fired'.''}, the prefix in configuration (2) would be \textbf{``This is a Attack event.''}, and in configuration (3) there would be no prefix.
\begin{table}[t]
    \centering
    \begin{tabular}{ccc}
    \toprule
        Configurations & F1 & $\Delta$ \\
        \midrule
        complete prefix prompt & 66.1 & - \\ 
        \midrule
        w/o event type & 64.4 & -1.7 \\ 
        w/o trigger & 64.9 & -1.2 \\ 
        w/o prefix prompt & 62.8 & -3.3 \\ 
        \bottomrule
    \end{tabular}
    \caption{Results of using different configurations of prefix prompt.}
    \label{tab:prefix}
\end{table}
The corresponding results are shown in Table \ref{tab:prefix}, where we have the following observations. First, removing either event type or trigger from the prefix prompt will cause a performance drop, which indicates that both kinds of information have contributions to the prompting process. Second, event type plays a more significant role than trigger does in prefix prompt, and the joint effect of them is greater than the sum of their respective effects.

\begin{figure}[t]
    \centering
    \resizebox{\columnwidth}{!}{
    \includegraphics{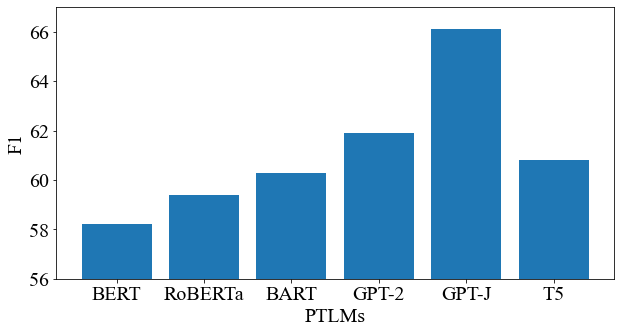}
    }
    \caption{Comparison between the performance of using different PTLMs in prompting module.}
    \label{fig:analysis_prompt_ptlm}
\end{figure}

In addition, we examine the effects of using different PTLMs in the prompting module.
We compare the following PTLMs with GPT-J (6B): BERT (large, uncased) \citep{DevlinCLT19}, RoBERTa (large) \citep{roberta}, BART (large) \citep{LewisLGGMLSZ20}, GPT-2 (xl) \citep{radford2019language}, T5 (11B) \citep{RaffelSRLNMZLL20}. The results are shown in Figure \ref{fig:analysis_prompt_ptlm}, where we have the following observations. 
First, the instance using GPT-J has the best performance, surpassing other instances by 4.2\% to 7.9\%. This shows that GPT-J has a better ability to understand events and their associated arguments compared to other PTLMs. Second, as PTLMs are listed in ascending order based on their numbers of parameters, we can see that for the first five models, the performance increases as the sizes of PTLMs become larger, which is consistent with the widely accepted notion that the larger model has a better capability of solving language tasks. However, the instance using the largest PTLM, T5 (11B), has a worse performance than GPT-2 and GPT-J. This is probably because autoregressive language modeling is more suitable for capturing information related to event arguments than mask language modeling is.

\subsection{Analysis of Global Constraints Regularization Module}\label{sec:analyze_constraint}
We conduct experiments to study the individual effect of each global constraint on the overall performance.
The results are shown in Table \ref{tab:used_constraints}, where we have the following observations. 
\begin{table}[!h]
    \centering
    \begin{tabular}{cccc}
    \toprule
        ~ & Model & F1 & $\Delta$ \\ 
        \midrule
        ~ & Full model & 66.1 & - \\ 
        \midrule
        ~ & w/o cross-task constraint & 60.5 & -5.6 \\ 
        ~ & w/o cross-argument constraint & 64.8 & -1.3 \\ 
        ~ & w/o cross-event constraint & 63.6 & -2.5 \\ 
        \bottomrule
    \end{tabular}
    \caption{Results of using different configurations of global constraints.}
    \label{tab:used_constraints}
\end{table}
First, every global constraint used by our model is beneficial to overall performance, which demonstrates that exploiting the domain knowledge about cross-task, cross-argument, and cross-event relations indeed provides our model with global understanding of event arguments. Second, the contribution of cross-task constraint is the most significant, which suggests that the global insights from the entity typing tasks are more effective in improving our model's reasoning ability about event arguments. Third, the cross-argument constraint is less effective than the other constraints, which shows that the global insights provided by the cross-argument constraint is less informative than those provided by the other constraints. 

Apart from the three global constraints described above, we have designed another 11 global constraints, which rely on cross-argument or cross-event relations. We add each of them into our model to check their respective effects on the overall performance.
The results of three of them are in Table \ref{tab:other_constraints}, whereas the results of all of them are in Section \ref{sec:comp_other_cons}. 
From the results, we can find that each of these constraints either brings minor improvement or even has a negative influence on the overall performance. 
Hence, we do not incorporate these constraints in our model to maintain our model's efficiency and effectiveness. %
\begin{table*}[t]
    \centering
    \resizebox{2\columnwidth}{!}{
    \begin{tabular}{cc}
    \toprule
    Global constraint & Effect on overall performance \\ 
    \midrule
    There is at most one Time-Arg in each event. & 0.4  \\ 
    A TRANSPORT event has at most one ORIGIN argument. & -0.1 \\ 
    If an Arrest-Jail event and a Charge-Indict event share arguments, & \multirow{3}{*}{0.3} \\
    Arrest-Jail.Person is the same as Charge-Indict.Defendant, they & \\
    share the same Crime argument. & \\ 
    \bottomrule
    \end{tabular}
    }
    \caption{Results of three other global constraints. Results of all other global constraints are in Section \ref{sec:comp_other_cons}}
    \label{tab:other_constraints}
\end{table*}

\subsection{Error Analysis}\label{sec:error_analysis}
\begin{figure}[t]
    \centering
    \resizebox{0.85\columnwidth}{!}{
    \includegraphics{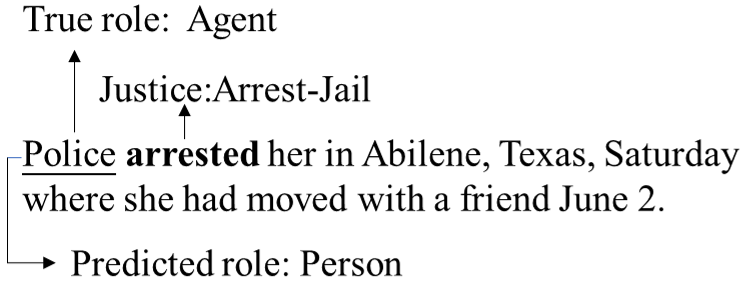}
    }
    \caption{An Example of the wrong prediction caused by too general argument roles. The text in \textbf{bold face} denotes trigger and the \underline{underlined text} denotes target argument span.}
    \label{fig:error_analysis}
\end{figure}

We manually checked 100 wrong predictions of our model and found that most of the errors are caused by too general roles of some event types. Specifically, some roles' linguistic meanings are so general that a model, not knowing their detailed event-type-dependent semantics, tends to assign them to some arguments which should have been assigned other roles. An example is shown in Figure \ref{fig:error_analysis}.
The example describes a Justice:Arrest-Jail event, which is associated with the following roles: ``Person,'' ``Agent,'' ``Crime,'' ``Time,'' and ``Place.''  ``Person'' refers to the person who is jailed or arrested, whereas ``Agent'' refers to the jailer or the arresting agent. In the example, the argument span's true role should be ``Agent'' according to the detailed event-type-dependent semantics of ``Person'' and ``Agent.''
However, our approach is zero-shot and directly models all role labels as natural language words, without incorporating the detailed event-type-dependent semantics of those roles, which are too general (e.g., ``Person''). 
Therefore, our model assigns ``Person'' to ``Police'' since it is reasonable from the perspectives of linguistic and commonsense knowledge, and ``Person'' is much more common than ``Agent'' in the pre-training corpus of the PTLM in the prompting module, which makes it have much higher likelihood in the language modeling process.
Incorporating event-type-dependent semantics of the roles which are too general into our model is left as future work.

\section{Related Work}
In this section, we introduce related works about constraint modeling, event extractions, and prompt-based Information Extraction (IE).

\subsection{Constraint Modeling}
Constraint modeling, as an important technique in machine learning and NLP, aims to improve a model's performance by incorporating domain knowledge as constraints~\cite{ganchev2010posterior,chang2012structured,ChangSR13,DeutschUR19,ChangRRR08,CGRS10,GracaGT10}.
One of the most significant advantages of constrained modeling is that it enables a model to capture the expressive and complex dependency structure in structured prediction problems like EAC~\cite{chang2012structured}.
Especially in zero-shot scenarios, constrained modeling can provide useful indirect supervision to a model, which further boosts performance~\cite{ganchev2010posterior}.
Some previous works have adopted constraints based on event-related domain knowledge to classify event arguments~\cite{LinJHW20,ZhangWR21}.
However, their constraints either require labor-intensive annotations~\cite{LinJHW20} or consider limited global information (e.g., cross-event relations)~\cite{ZhangWR21}.
In this paper, our model uses global constraints to regularize prediction by incorporating global insights from cross-task, cross-argument, and cross-event relations.

\subsection{Event Extraction}

Event extraction is a fundamental information extraction task~\cite{DBLP:conf/muc/Sundheim92,DBLP:conf/coling/GrishmanS96,DBLP:conf/aaai/Riloff96,grishman2005nyu,chen2021event,DuC20,LiuCLBL20}, which can be further divided into four sub-tasks: trigger identification, trigger classification, argument identification, and argument classification.
Traditional efforts mostly focus on the supervised setting~\cite{JiG08,DBLP:conf/acl/LiaoG10,DBLP:conf/acl/LiuCHL016,DBLP:conf/acl/ChenXLZ015,DBLP:conf/naacl/NguyenCG16,DBLP:conf/emnlp/LiuLH18,DBLP:journals/dint/ZhangJS19,DBLP:conf/emnlp/WaddenWLH19,LinJHW20}.
However, these works could suffer from the huge burden of human annotation.
In this work, we focus on the argument classification task and propose a model using prompting and global constraints, without annotation and task-specific training.

\subsection{Prompt-based IE}

With the fast development of large PTLMs like T5~\cite{RaffelSRLNMZLL20}, GPT-3~\cite{DBLP:conf/nips/BrownMRSKDNSSAA20}, and Pathway Language models~\cite{DBLP:journals/corr/abs-2204-02311}, the prompt-based method has been an efficient tool of applying those giant models into downstream NLP tasks~\cite{DBLP:journals/corr/abs-2107-13586}. IE is not an exception.
People have been using leverage prompts and giant models to solve IE tasks like named entity recognition~\cite{DBLP:conf/acl/CuiWLYZ21}, semantic parsing~\cite{DBLP:conf/emnlp/ShinLTCRPPKED21}, and relations extraction~\cite{DBLP:journals/corr/abs-2202-04824,DBLP:journals/corr/abs-2105-11259} in a zero-shot or few-shot way.
However, previous prompting methods for IE need a tedious prompt design for every new type of events and arguments.
In contrast, our model's prompt templates can be adapted to all possible types of events and arguments in a fully automatic way.

\section{Conclusion}
We propose a zero-shot EAC model using global constraints with prompting.
Compared with previous works, our model does not require any annotation or manual prompt design, and our constraint modeling method can be easily adapted to any other datasets.
Hence, our model can be easily generalized to any open-world event ontologies.
Experiments on two standard event extraction datasets demonstrate our model's effectiveness.

\section{Limitations}
Our work has the following limitations. 
One limitation is that our model is not aware of the detailed event-type-dependent semantics of those roles which are too general, as discussed in Section 3.5. In the future, we will work on enabling our model to capture the event-type-dependent semantics of the roles which are too general.
Another limitation is that our model's performance is still unsatisfactory compared with SOTA supervised model when argument spans are not given, as discussed in Section 3.2. 
In the future, we will work on designing a more powerful zero-shot event argument identification module for our model, so that we can obtain satisfactory zero-shot EAC performance even when argument spans are not given.

\section{Acknowledgement}
The authors of this paper were supported by the NSFC Fund (U20B2053) from the NSFC of China, the RIF (R6020-19 and R6021-20) and the GRF (16211520 and 16205322) from RGC of Hong Kong, the MHKJFS (MHP/001/19) from ITC of Hong Kong and the National Key R\&D Program of China (2019YFE0198200) with special thanks to HKMAAC and CUSBLT, and the Jiangsu Province Science and Technology Collaboration Fund (BZ2021065). We also thank the support from the UGC Research Matching Grants (RMGS20EG01-D, RMGS20CR11, RMGS20CR12, RMGS20EG19, RMGS20EG21, RMGS23CR05, RMGS23EG08).

\bibliographystyle{acl_natbib}
\bibliography{custom}

\begin{thebibliography}{51}
\expandafter\ifx\csname natexlab\endcsname\relax\def\natexlab#1{#1}\fi

\bibitem[{Brown et~al.(2020)Brown, Mann, Ryder, Subbiah, Kaplan, Dhariwal,
  Neelakantan, Shyam, Sastry, Askell, Agarwal, Herbert{-}Voss, Krueger,
  Henighan, Child, Ramesh, Ziegler, Wu, Winter, Hesse, Chen, Sigler, Litwin,
  Gray, Chess, Clark, Berner, McCandlish, Radford, Sutskever, and
  Amodei}]{DBLP:conf/nips/BrownMRSKDNSSAA20}
Tom~B. Brown, Benjamin Mann, Nick Ryder, Melanie Subbiah, Jared Kaplan,
  Prafulla Dhariwal, Arvind Neelakantan, Pranav Shyam, Girish Sastry, Amanda
  Askell, Sandhini Agarwal, Ariel Herbert{-}Voss, Gretchen Krueger, Tom
  Henighan, Rewon Child, Aditya Ramesh, Daniel~M. Ziegler, Jeffrey Wu, Clemens
  Winter, Christopher Hesse, Mark Chen, Eric Sigler, Mateusz Litwin, Scott
  Gray, Benjamin Chess, Jack Clark, Christopher Berner, Sam McCandlish, Alec
  Radford, Ilya Sutskever, and Dario Amodei. 2020.
\newblock Language models are few-shot learners.
\newblock In \emph{NeurIPS}.

\bibitem[{Chang et~al.(2013)Chang, Samdani, and Roth}]{ChangSR13}
Kai{-}Wei Chang, Rajhans Samdani, and Dan Roth. 2013.
\newblock A constrained latent variable model for coreference resolution.
\newblock In \emph{EMNLP}, pages 601--612. {ACL}.

\bibitem[{Chang et~al.(2010)Chang, Goldwasser, Roth, and Srikumar}]{CGRS10}
Ming-Wei Chang, Dan Goldwasser, Dan Roth, and Vivek Srikumar. 2010.
\newblock {Discriminative Learning over Constrained Latent Representations}.
\newblock In \emph{NAACL}. ACL.

\bibitem[{Chang et~al.(2012)Chang, Ratinov, and Roth}]{chang2012structured}
Ming-Wei Chang, Lev Ratinov, and Dan Roth. 2012.
\newblock Structured learning with constrained conditional models.
\newblock \emph{Machine learning}, 88(3):399--431.

\bibitem[{Chang et~al.(2008)Chang, Ratinov, Rizzolo, and Roth}]{ChangRRR08}
Ming{-}Wei Chang, Lev{-}Arie Ratinov, Nicholas Rizzolo, and Dan Roth. 2008.
\newblock Learning and inference with constraints.
\newblock In \emph{AAAI}, pages 1513--1518. AAAI Press.

\bibitem[{Chen et~al.(2021)Chen, Zhang, Ning, Li, Ji, McKeown, and
  Roth}]{chen2021event}
Muhao Chen, Hongming Zhang, Qiang Ning, Manling Li, Heng Ji, Kathleen McKeown,
  and Dan Roth. 2021.
\newblock Event-centric natural language processing.
\newblock In \emph{ACL Tutorial}, pages 6--14.

\bibitem[{Chen et~al.(2015)Chen, Xu, Liu, Zeng, and
  Zhao}]{DBLP:conf/acl/ChenXLZ015}
Yubo Chen, Liheng Xu, Kang Liu, Daojian Zeng, and Jun Zhao. 2015.
\newblock \href {https://doi.org/10.3115/v1/p15-1017} {Event extraction via
  dynamic multi-pooling convolutional neural networks}.
\newblock In \emph{ACL}, pages 167--176. ACL.

\bibitem[{Chen et~al.(2022)Chen, Liu, Dong, Wang, Zhu, Zeng, and
  Zhang}]{DBLP:journals/corr/abs-2202-04824}
Yulong Chen, Yang Liu, Li~Dong, Shuohang Wang, Chenguang Zhu, Michael Zeng, and
  Yue Zhang. 2022.
\newblock Adaprompt: Adaptive model training for prompt-based {NLP}.
\newblock \emph{CoRR}, abs/2202.04824.

\bibitem[{Chowdhery et~al.(2022)Chowdhery, Narang, Devlin, Bosma, Mishra,
  Roberts, Barham, Chung, Sutton, Gehrmann, Schuh, Shi, Tsvyashchenko, Maynez,
  Rao, Barnes, Tay, Shazeer, Prabhakaran, Reif, Du, Hutchinson, Pope, Bradbury,
  Austin, Isard, Gur{-}Ari, Yin, Duke, Levskaya, Ghemawat, Dev, Michalewski,
  Garcia, Misra, Robinson, Fedus, Zhou, Ippolito, Luan, Lim, Zoph, Spiridonov,
  Sepassi, Dohan, Agrawal, Omernick, Dai, Pillai, Pellat, Lewkowycz, Moreira,
  Child, Polozov, Lee, Zhou, Wang, Saeta, Diaz, Firat, Catasta, Wei,
  Meier{-}Hellstern, Eck, Dean, Petrov, and
  Fiedel}]{DBLP:journals/corr/abs-2204-02311}
Aakanksha Chowdhery, Sharan Narang, Jacob Devlin, Maarten Bosma, Gaurav Mishra,
  Adam Roberts, Paul Barham, Hyung~Won Chung, Charles Sutton, Sebastian
  Gehrmann, Parker Schuh, Kensen Shi, Sasha Tsvyashchenko, Joshua Maynez,
  Abhishek Rao, Parker Barnes, Yi~Tay, Noam Shazeer, Vinodkumar Prabhakaran,
  Emily Reif, Nan Du, Ben Hutchinson, Reiner Pope, James Bradbury, Jacob
  Austin, Michael Isard, Guy Gur{-}Ari, Pengcheng Yin, Toju Duke, Anselm
  Levskaya, Sanjay Ghemawat, Sunipa Dev, Henryk Michalewski, Xavier Garcia,
  Vedant Misra, Kevin Robinson, Liam Fedus, Denny Zhou, Daphne Ippolito, David
  Luan, Hyeontaek Lim, Barret Zoph, Alexander Spiridonov, Ryan Sepassi, David
  Dohan, Shivani Agrawal, Mark Omernick, Andrew~M. Dai,
  Thanumalayan~Sankaranarayana Pillai, Marie Pellat, Aitor Lewkowycz, Erica
  Moreira, Rewon Child, Oleksandr Polozov, Katherine Lee, Zongwei Zhou, Xuezhi
  Wang, Brennan Saeta, Mark Diaz, Orhan Firat, Michele Catasta, Jason Wei,
  Kathy Meier{-}Hellstern, Douglas Eck, Jeff Dean, Slav Petrov, and Noah
  Fiedel. 2022.
\newblock Palm: Scaling language modeling with pathways.
\newblock \emph{CoRR}, abs/2204.02311.

\bibitem[{Cui et~al.(2021)Cui, Wu, Liu, Yang, and
  Zhang}]{DBLP:conf/acl/CuiWLYZ21}
Leyang Cui, Yu~Wu, Jian Liu, Sen Yang, and Yue Zhang. 2021.
\newblock Template-based named entity recognition using {BART}.
\newblock In \emph{Findings ofACL/IJCNLP}, pages 1835--1845. ACL.

\bibitem[{Dai et~al.(2021)Dai, Song, and Wang}]{DaiSW20}
Hongliang Dai, Yangqiu Song, and Haixun Wang. 2021.
\newblock Ultra-fine entity typing with weak supervision from a masked language
  model.
\newblock In \emph{ACL/IJCNLP}, pages 1790--1799. ACL.

\bibitem[{Deutsch et~al.(2019)Deutsch, Upadhyay, and Roth}]{DeutschUR19}
Daniel Deutsch, Shyam Upadhyay, and Dan Roth. 2019.
\newblock A general-purpose algorithm for constrained sequential inference.
\newblock In \emph{CoNLL}, pages 482--492. ACL.

\bibitem[{Devlin et~al.(2019)Devlin, Chang, Lee, and Toutanova}]{DevlinCLT19}
Jacob Devlin, Ming{-}Wei Chang, Kenton Lee, and Kristina Toutanova. 2019.
\newblock {BERT:} pre-training of deep bidirectional transformers for language
  understanding.
\newblock In \emph{NAACL}, pages 4171--4186. ACL.

\bibitem[{Ding et~al.(2015)Ding, Zhang, Liu, and Duan}]{DingZLD15}
Xiao Ding, Yue Zhang, Ting Liu, and Junwen Duan. 2015.
\newblock Deep learning for event-driven stock prediction.
\newblock In \emph{IJCAI}, pages 2327--2333. AAAI Press.

\bibitem[{Doddington et~al.(2004)Doddington, Mitchell, Przybocki, Ramshaw,
  Strassel, and Weischedel}]{DoddingtonMPRSW04}
George~R. Doddington, Alexis Mitchell, Mark~A. Przybocki, Lance~A. Ramshaw,
  Stephanie~M. Strassel, and Ralph~M. Weischedel. 2004.
\newblock The automatic content extraction {(ACE)} program - tasks, data, and
  evaluation.
\newblock In \emph{LREC}. ELRA.

\bibitem[{Du and Cardie(2020)}]{DuC20}
Xinya Du and Claire Cardie. 2020.
\newblock \href {https://doi.org/10.18653/v1/2020.emnlp-main.49} {Event
  extraction by answering (almost) natural questions}.
\newblock In \emph{Proceedings of the 2020 Conference on Empirical Methods in
  Natural Language Processing, {EMNLP} 2020, Online, November 16-20, 2020},
  pages 671--683. Association for Computational Linguistics.

\bibitem[{Ganchev et~al.(2010)Ganchev, Gra{\~A}, Gillenwater, and
  Taskar}]{ganchev2010posterior}
Kuzman Ganchev, Jo{\~a}o Gra{\~A}, Jennifer Gillenwater, and Ben Taskar. 2010.
\newblock Posterior regularization for structured latent variable models.
\newblock \emph{Journal of Machine Learning Research}, 11(67):2001--2049.

\bibitem[{Gra{\c{c}}a et~al.(2010)Gra{\c{c}}a, Ganchev, and Taskar}]{GracaGT10}
Jo{\~{a}}o Gra{\c{c}}a, Kuzman Ganchev, and Ben Taskar. 2010.
\newblock Learning tractable word alignment models with complex constraints.
\newblock \emph{Comput. Linguistics}, 36(3):481--504.

\bibitem[{Grishman and Sundheim(1996)}]{DBLP:conf/coling/GrishmanS96}
Ralph Grishman and Beth Sundheim. 1996.
\newblock Message understanding conference- 6: {A} brief history.
\newblock In \emph{COLING}, pages 466--471.

\bibitem[{Grishman et~al.(2005)Grishman, Westbrook, and
  Meyers}]{grishman2005nyu}
Ralph Grishman, David Westbrook, and Adam Meyers. 2005.
\newblock Nyu’s english ace 2005 system description.
\newblock \emph{ACE}, 5.

\bibitem[{Han et~al.(2021)Han, Zhao, Ding, Liu, and
  Sun}]{DBLP:journals/corr/abs-2105-11259}
Xu~Han, Weilin Zhao, Ning Ding, Zhiyuan Liu, and Maosong Sun. 2021.
\newblock {PTR:} prompt tuning with rules for text classification.
\newblock \emph{CoRR}, abs/2105.11259.

\bibitem[{Hsu et~al.(2022)Hsu, Huang, Boschee, Miller, Natarajan, Chang, and
  Peng}]{naacl2022degree}
I-Hung Hsu, Kuan-Hao Huang, Elizabeth Boschee, Scott Miller, Prem Natarajan,
  Kai-Wei Chang, and Nanyun Peng. 2022.
\newblock Degree: A data-efficient generative event extraction model.
\newblock In \emph{NAACL}. ACL.

\bibitem[{Huang et~al.(2022)Huang, Hsu, Natarajan, Chang, and
  Peng}]{HuangHNCP22}
Kuan{-}Hao Huang, I{-}Hung Hsu, Prem Natarajan, Kai{-}Wei Chang, and Nanyun
  Peng. 2022.
\newblock Multilingual generative language models for zero-shot cross-lingual
  event argument extraction.
\newblock In \emph{ACL}, pages 4633--4646. ACL.

\bibitem[{Huang et~al.(2018)Huang, Ji, Cho, Dagan, Riedel, and
  Voss}]{DaganJVHCR18}
Lifu Huang, Heng Ji, Kyunghyun Cho, Ido Dagan, Sebastian Riedel, and Clare~R.
  Voss. 2018.
\newblock Zero-shot transfer learning for event extraction.
\newblock In \emph{ACL}, pages 2160--2170. ACL.

\bibitem[{Ji and Grishman(2008)}]{JiG08}
Heng Ji and Ralph Grishman. 2008.
\newblock Refining event extraction through cross-document inference.
\newblock In \emph{ACL}, pages 254--262. ACL.

\bibitem[{Lewis et~al.(2020)Lewis, Liu, Goyal, Ghazvininejad, Mohamed, Levy,
  Stoyanov, and Zettlemoyer}]{LewisLGGMLSZ20}
Mike Lewis, Yinhan Liu, Naman Goyal, Marjan Ghazvininejad, Abdelrahman Mohamed,
  Omer Levy, Veselin Stoyanov, and Luke Zettlemoyer. 2020.
\newblock {BART:} denoising sequence-to-sequence pre-training for natural
  language generation, translation, and comprehension.
\newblock In \emph{ACL}, pages 7871--7880. ACL.

\bibitem[{Liao and Grishman(2010)}]{DBLP:conf/acl/LiaoG10}
Shasha Liao and Ralph Grishman. 2010.
\newblock Using document level cross-event inference to improve event
  extraction.
\newblock In \emph{ACL}, pages 789--797. ACL.

\bibitem[{Lin et~al.(2020)Lin, Ji, Huang, and Wu}]{LinJHW20}
Ying Lin, Heng Ji, Fei Huang, and Lingfei Wu. 2020.
\newblock A joint neural model for information extraction with global features.
\newblock In \emph{ACL}, pages 7999--8009. ACL.

\bibitem[{Liu et~al.(2020)Liu, Chen, Liu, Bi, and Liu}]{LiuCLBL20}
Jian Liu, Yubo Chen, Kang Liu, Wei Bi, and Xiaojiang Liu. 2020.
\newblock Event extraction as machine reading comprehension.
\newblock In \emph{EMNLP}, pages 1641--1651. ACL.

\bibitem[{Liu et~al.(2021)Liu, Yuan, Fu, Jiang, Hayashi, and
  Neubig}]{DBLP:journals/corr/abs-2107-13586}
Pengfei Liu, Weizhe Yuan, Jinlan Fu, Zhengbao Jiang, Hiroaki Hayashi, and
  Graham Neubig. 2021.
\newblock Pre-train, prompt, and predict: {A} systematic survey of prompting
  methods in natural language processing.
\newblock \emph{CoRR}, abs/2107.13586.

\bibitem[{Liu et~al.(2016)Liu, Chen, He, Liu, and
  Zhao}]{DBLP:conf/acl/LiuCHL016}
Shulin Liu, Yubo Chen, Shizhu He, Kang Liu, and Jun Zhao. 2016.
\newblock Leveraging framenet to improve automatic event detection.
\newblock In \emph{ACL}. ACL.

\bibitem[{Liu et~al.(2022)Liu, Huang, Shi, and Wang}]{LiuHSW22}
Xiao Liu, Heyan Huang, Ge~Shi, and Bo~Wang. 2022.
\newblock \href {https://aclanthology.org/2022.acl-long.358} {Dynamic
  prefix-tuning for generative template-based event extraction}.
\newblock In \emph{ACL}, pages 5216--5228. ACL.

\bibitem[{Liu et~al.(2018)Liu, Luo, and Huang}]{DBLP:conf/emnlp/LiuLH18}
Xiao Liu, Zhunchen Luo, and Heyan Huang. 2018.
\newblock Jointly multiple events extraction via attention-based graph
  information aggregation.
\newblock In \emph{EMNLP}, pages 1247--1256.

\bibitem[{Liu et~al.(2019)Liu, Ott, Goyal, Du, Joshi, Chen, Levy, Lewis,
  Zettlemoyer, and Stoyanov}]{roberta}
Yinhan Liu, Myle Ott, Naman Goyal, Jingfei Du, Mandar Joshi, Danqi Chen, Omer
  Levy, Mike Lewis, Luke Zettlemoyer, and Veselin Stoyanov. 2019.
\newblock Roberta: {A} robustly optimized {BERT} pretraining approach.
\newblock \emph{CoRR}, abs/1907.11692.

\bibitem[{Lyu et~al.(2021)Lyu, Zhang, Sulem, and Roth}]{LyuZSR20}
Qing Lyu, Hongming Zhang, Elior Sulem, and Dan Roth. 2021.
\newblock Zero-shot event extraction via transfer learning: Challenges and
  insights.
\newblock In \emph{ACL/IJCNLP (Short Papers)}, pages 322--332. ACL.

\bibitem[{Ma et~al.(2022)Ma, Wang, Cao, Li, Chen, Wang, and Shao}]{MaW0LCWS22}
Yubo Ma, Zehao Wang, Yixin Cao, Mukai Li, Meiqi Chen, Kun Wang, and Jing Shao.
  2022.
\newblock Prompt for extraction? {PAIE:} prompting argument interaction for
  event argument extraction.
\newblock In \emph{ACL}, pages 6759--6774. ACL.

\bibitem[{Mehta et~al.(2022)Mehta, Rangwala, and Ramakrishnan}]{abs-2204-02531}
Sneha Mehta, Huzefa Rangwala, and Naren Ramakrishnan. 2022.
\newblock \href {https://doi.org/10.48550/arXiv.2204.02531} {Improving
  zero-shot event extraction via sentence simplification}.
\newblock \emph{CoRR}, abs/2204.02531.

\bibitem[{Nguyen et~al.(2016)Nguyen, Cho, and
  Grishman}]{DBLP:conf/naacl/NguyenCG16}
Thien~Huu Nguyen, Kyunghyun Cho, and Ralph Grishman. 2016.
\newblock Joint event extraction via recurrent neural networks.
\newblock In \emph{NAACL-HLT}, pages 300--309. ACL.

\bibitem[{Radford et~al.(2019)Radford, Wu, Child, Luan, Amodei, and
  Sutskever}]{radford2019language}
Alec Radford, Jeff Wu, Rewon Child, David Luan, Dario Amodei, and Ilya
  Sutskever. 2019.
\newblock Language models are unsupervised multitask learners.

\bibitem[{Raffel et~al.(2020)Raffel, Shazeer, Roberts, Lee, Narang, Matena,
  Zhou, Li, and Liu}]{RaffelSRLNMZLL20}
Colin Raffel, Noam Shazeer, Adam Roberts, Katherine Lee, Sharan Narang, Michael
  Matena, Yanqi Zhou, Wei Li, and Peter~J. Liu. 2020.
\newblock Exploring the limits of transfer learning with a unified text-to-text
  transformer.
\newblock \emph{J. Mach. Learn. Res.}, 21:140:1--140:67.

\bibitem[{Riloff(1996)}]{DBLP:conf/aaai/Riloff96}
Ellen Riloff. 1996.
\newblock Automatically generating extraction patterns from untagged text.
\newblock In \emph{AAAI}, pages 1044--1049. AAAI Press.

\bibitem[{Shin et~al.(2021)Shin, Lin, Thomson, Chen, Roy, Platanios, Pauls,
  Klein, Eisner, and Durme}]{DBLP:conf/emnlp/ShinLTCRPPKED21}
Richard Shin, Christopher~H. Lin, Sam Thomson, Charles Chen, Subhro Roy,
  Emmanouil~Antonios Platanios, Adam Pauls, Dan Klein, Jason Eisner, and
  Benjamin~Van Durme. 2021.
\newblock Constrained language models yield few-shot semantic parsers.
\newblock In \emph{EMNLP}, pages 7699--7715. ACL.

\bibitem[{Song et~al.(2015)Song, Bies, Strassel, Riese, Mott, Ellis, Wright,
  Kulick, Ryant, and Ma}]{SongBSRMEWKRM15}
Zhiyi Song, Ann Bies, Stephanie~M. Strassel, Tom Riese, Justin Mott, Joe Ellis,
  Jonathan Wright, Seth Kulick, Neville Ryant, and Xiaoyi Ma. 2015.
\newblock From light to rich {ERE:} annotation of entities, relations, and
  events.
\newblock In \emph{The 3rd Workshop on {EVENTS:} Definition, Detection,
  Coreference, and Representation, EVENTS@HLP-NAACL}, pages 89--98. ACL.

\bibitem[{Sundheim(1992)}]{DBLP:conf/muc/Sundheim92}
Beth Sundheim. 1992.
\newblock Overview of the fourth message understanding evaluation and
  conference.
\newblock In \emph{MUC}, pages 3--21.

\bibitem[{Wadden et~al.(2019)Wadden, Wennberg, Luan, and
  Hajishirzi}]{DBLP:conf/emnlp/WaddenWLH19}
David Wadden, Ulme Wennberg, Yi~Luan, and Hannaneh Hajishirzi. 2019.
\newblock Entity, relation, and event extraction with contextualized span
  representations.
\newblock In \emph{EMNLP-IJCNLP}, pages 5783--5788. ACL.

\bibitem[{Wang and Komatsuzaki(2021)}]{gpt-j}
Ben Wang and Aran Komatsuzaki. 2021.
\newblock {GPT-J-6B: A 6 Billion Parameter Autoregressive Language Model}.
\newblock \url{https://github.com/kingoflolz/mesh-transformer-jax}.

\bibitem[{Wang et~al.(2022)Wang, Yu, Chang, Sun, and Huang}]{WangYCSH22}
Sijia Wang, Mo~Yu, Shiyu Chang, Lichao Sun, and Lifu Huang. 2022.
\newblock Query and extract: Refining event extraction as type-oriented binary
  decoding.
\newblock In \emph{Findings of ACL}, pages 169--182. ACL.

\bibitem[{Wolf et~al.(2020)Wolf, Debut, Sanh, Chaumond, Delangue, Moi, Cistac,
  Rault, Louf, Funtowicz, Davison, Shleifer, von Platen, Ma, Jernite, Plu, Xu,
  Scao, Gugger, Drame, Lhoest, and Rush}]{WolfDSCDMCRLFDS20}
Thomas Wolf, Lysandre Debut, Victor Sanh, Julien Chaumond, Clement Delangue,
  Anthony Moi, Pierric Cistac, Tim Rault, R{\'{e}}mi Louf, Morgan Funtowicz,
  Joe Davison, Sam Shleifer, Patrick von Platen, Clara Ma, Yacine Jernite,
  Julien Plu, Canwen Xu, Teven~Le Scao, Sylvain Gugger, Mariama Drame, Quentin
  Lhoest, and Alexander~M. Rush. 2020.
\newblock Transformers: State-of-the-art natural language processing.
\newblock In \emph{EMNLP (Demos)}, pages 38--45. ACL.

\bibitem[{Zhang et~al.(2021)Zhang, Wang, and Roth}]{ZhangWR21}
Hongming Zhang, Haoyu Wang, and Dan Roth. 2021.
\newblock Zero-shot label-aware event trigger and argument classification.
\newblock In \emph{Findings of ACL/IJCNLP}, pages 1331--1340. ACL.

\bibitem[{Zhang et~al.(2019)Zhang, Ji, and Sil}]{DBLP:journals/dint/ZhangJS19}
Tongtao Zhang, Heng Ji, and Avirup Sil. 2019.
\newblock Joint entity and event extraction with generative adversarial
  imitation learning.
\newblock \emph{Data Intell.}, 1(2):99--120.

\bibitem[{Zhao et~al.(2021)Zhao, Zhang, Yang, He, Ma, and Li}]{ZhaoZYHML21}
Weizhong Zhao, Jinyong Zhang, Jincai Yang, Tingting He, Huifang Ma, and Zhixin
  Li. 2021.
\newblock A novel joint biomedical event extraction framework via two-level
  modeling of documents.
\newblock \emph{Inf. Sci.}, 550:27--40.

\end{thebibliography}

\appendix
\section{Comparison between Different Prefix Prompts}\label{sec:prefix_comp}
In this section, we conduct experiments on ACE-2005 dataset to compare the effectiveness of using different prefix prompts in our models. We compare the following prefix prompts with the one discussed in Section \ref{sec:prompting_module}: (1) ``\textbf{This is a [] event whose trigger is "[]".}''; (2) ``\textbf{The event type is [], and its occurrence is most clearly expressed by "[]".}''; (3) ``\textbf{The event type is [] and the trigger is "[]".}''. The results are shown in Table \ref{tab:diff_prefix}, where ``Prefix (0)'' refers to the prefix prompt discussed in Section \ref{sec:prompting_module}, whereas ``Prefix (1)'' refers to the first prefix prompt described in this section, and so on.
\begin{table}[!h]
    \centering
    \begin{tabular}{cc}
    \toprule
        Prefix Prompt & F1\\ 
        \midrule
        Prefix (0) & \textbf{66.1}\\ 
        Prefix (1) & 65.2\\ 
        Prefix (2) & 65.6\\ 
        Prefix (3) & 63.0\\ 
        \bottomrule
    \end{tabular}
    \caption{Performance of different prefix prompts.}
    \label{tab:diff_prefix}
\end{table}
From the table we can see that the prefix prompt described in Section \ref{sec:prompting_module} is the most effective one, which might be due to the fact that the prefix prompt not only is based on the definitions of events and triggers \citep{grishman2005nyu}, but also has a natural and smooth expression. 

\section{Results of all Other Global Constraints}\label{sec:comp_other_cons}
\begin{table*}[!h]
    \centering
    \resizebox{2\columnwidth}{!}{
    \begin{tabular}{cc}
    \toprule
    Global constraint & Effect on overall performance \\ 
    \midrule
    There is at most one Time-Arg in each event. & 0.4  \\ 
    There is at most one Place-Arg in each event. & 0.1 \\ 
    A TRANSPORT event has at most one Destination argument. & -0.2 \\ 
    A TRANSPORT event has at most one ORIGIN argument. & -0.1 \\ 
    A START-POSITION event has at most one Person argument. & 0.2 \\ 
    A START-POSITION event has at most one Entity argument. & -0.1 \\ 
    A START-POSITION event has at most one Position argument. & 0.1 \\ 
    A End-POSITION event has at most one Person argument. & -0.2 \\ 
    If a Start-Position event and an End-Position event share & \multirow{5}{*}{
    0.1} \\
    arguments, then Start-Position.Person is the same as & \\
    End-Position.Person, and Start-Position.Entity is the same & \\
    as End-Position.Entity, Start-Position.Position is the same & \\
    as End-Position.Position. & \\ 
    If an Arrest-Jail event and a Charge-Indict event share arguments, & \multirow{3}{*}{0.3} \\
    Arrest-Jail.Person is the same as Charge-Indict.Defendant, they & \\
    share the same Crime argument. & \\ 
    If a Die event and an Attack event share arguments, then & \multirow{5}{*}{-0.2} \\
    Die.Place is the same as Attack.Place, Die.Victim is the &\\
    same as Attack.Target, Die.Instrument is the same as &\\
    Attack.Instrument, Die.Time is the same as Attack.Time, &\\
    Die.Agent is the same as Attack.Attacker. & \\ 
    \bottomrule
    \end{tabular}
    }
    \caption{Other global constraints and corresponding effects on overall performance.}
    \label{tab:comp_other_cons}
\end{table*}

In this section, we present the results of all other global constraints. The results are shown in Table \ref{tab:comp_other_cons}.

\end{document}